\documentclass[letterpaper, 10 pt, conference]{ieeeconf}

\IEEEoverridecommandlockouts
\usepackage{color}
\usepackage{amsfonts}
\usepackage{amsmath}
\usepackage{cite}
\usepackage{makecell}
\usepackage{siunitx}
\usepackage{gensymb}

\usepackage{array}
\newcolumntype{?}{!{\vrule width 2pt}}

\usepackage{booktabs}
\usepackage{multirow}

\usepackage{xcolor}

\definecolor{CommentTB}{rgb}{1.0,0,0} 
\definecolor{CommentUrgent}{rgb} {0.7,0.0,0.7} 
\definecolor{CommentHP}{rgb}{0.0,0.7,0.0} 
\definecolor{CommentLK}{rgb} {0,0.7,1.0} 

\newcommand{\tom}[1] {{\color{CommentTB} { Tom: \textbf{#1}}}}
\newcommand{\pmnnotes}[1] {{\color{CommentUrgent} { PaulN: \textbf{#1}}}}
\newcommand{\todo}[1] {{\color{CommentUrgent} { TODO: \textbf{#1}}}}

\newcommand{\ignore}[1]

\renewcommand{\pmnnotes}[1]{}
\renewcommand{\todo}[1]{}
\renewcommand{\ignore}[1]{}

  \usepackage[pdftex]{graphicx}
  \DeclareGraphicsExtensions{.pdf,.jpeg,.png}

\usepackage{tikz}
\usepackage{overpic}

\begin{document}

\title{The Right (Angled) Perspective: Improving the Understanding of Road Scenes Using Boosted Inverse Perspective Mapping}

\author{Tom Bruls$^\ast$, Horia Porav$^\ast$, Lars Kunze, and Paul Newman
\thanks{$\ast$ equal contribution} \thanks{{Authors are from the Oxford Robotics Institute, Dept. Engineering Science, University of Oxford, UK. \{\texttt{tombruls}, \texttt{horia}, \texttt{lars}, \texttt{pnewman}\}\texttt{@robots.ox.ac.uk}}}
}

\maketitle

\begin{abstract}
Many tasks performed by autonomous vehicles such as road marking detection, object tracking, and path planning are simpler in bird's-eye view.
Hence, Inverse Perspective Mapping (IPM) is often applied to remove the perspective effect from a vehicle's front-facing camera and to remap its images into a 2D domain, resulting in a top-down view.
Unfortunately, however, this leads to unnatural blurring and stretching of objects at further distance, due to the resolution of the camera, limiting applicability. 
In this paper, we present an adversarial learning approach for generating a significantly improved IPM from a single camera image in real time.
The generated bird's-eye-view images contain sharper features (e.g. road markings) and a more homogeneous illumination, while (dynamic) objects are automatically removed from the scene, thus revealing the underlying road layout in an improved fashion.
We demonstrate our framework using real-world data from the Oxford RobotCar Dataset and show that scene understanding tasks directly benefit from our boosted IPM approach.


\end{abstract}

\pmnnotes{OK this is shaping up nicely - it clearly works. It needs someone to go through and tighten up the language some word orders are unusual  - but I have to say with English as a second languauge you guy are incredible. But do get it checked. Ok, so, the biggest risk here is obvioulsy not having any quantitaive measures. We agreed that is really hard. SO we should be very explicit about that and say (ideally with pointers) that ground truthing overhead view synthesis is not possible because the no camera can physically be in the pose to acquire the image. So this is challenging. I think we should give up more of the lit review and instead break out LArs's sematic description figures into all their glory glory. If we spend text writing about taht we could show that the semantic decsription of the old-style IPM is nonsense. But this requires finding some serious space  - but to my mind it is worth it....   Lars  - you are a senior - can I leave this one with you to sign off? It obviously passes the quality threshold when touched up and edited.}

\section{Introduction}
\label{sec:introduction}

Autonomous vehicles need to perceive and fully understand their environment to accomplish their navigation tasks.
Hence, scene understanding is a critical component within their perception pipeline, not only for navigation and planning, but also for safety purposes.
While vehicles use different types of sensors to interpret scenes, cameras are one of the most popular sensing modalities in the field, due to their low cost as well as the availability of well-established image processing techniques.

In recent years, deep learning approaches based on images have been very successful and significantly improved the performance of autonomous vehicles in the context of semantic scene understanding~\cite{badrinarayanan2017segnet, schneider2016semantic}.
Many of these approaches take images from a front-facing camera as their input.
However, images as well as their interpretations (i.e. segmented pixels) in this perspective are often transformed into a local and/or global coordinate system (or view) to be utilized effectively within tasks such as lane detection~\cite{neven2018towards, song2018lane}, road marking detection~\cite{mathibela2015reading}, road topology detection~\cite{ballardini2017online, schulter2018learning}, object detection/tracking~\cite{dequaire2018deep, engel2018deep,simond2007obstacle}, as well as path planning and intersection prediction~\cite{lee2017desire, zyner2018naturalistic}.
This transformation is commonly referred to as Inverse Perspective Mapping (IPM) \cite{hartley2004multiple}.
IPM takes the frontal view as input, applies a homography, and produces a top-down view of the scene by mapping the pixels to a different 2D-coordinate frame, which is also known as \emph{bird's-eye view}.

\begin{figure}[!t]
	\centering
	\includegraphics[width=\columnwidth]{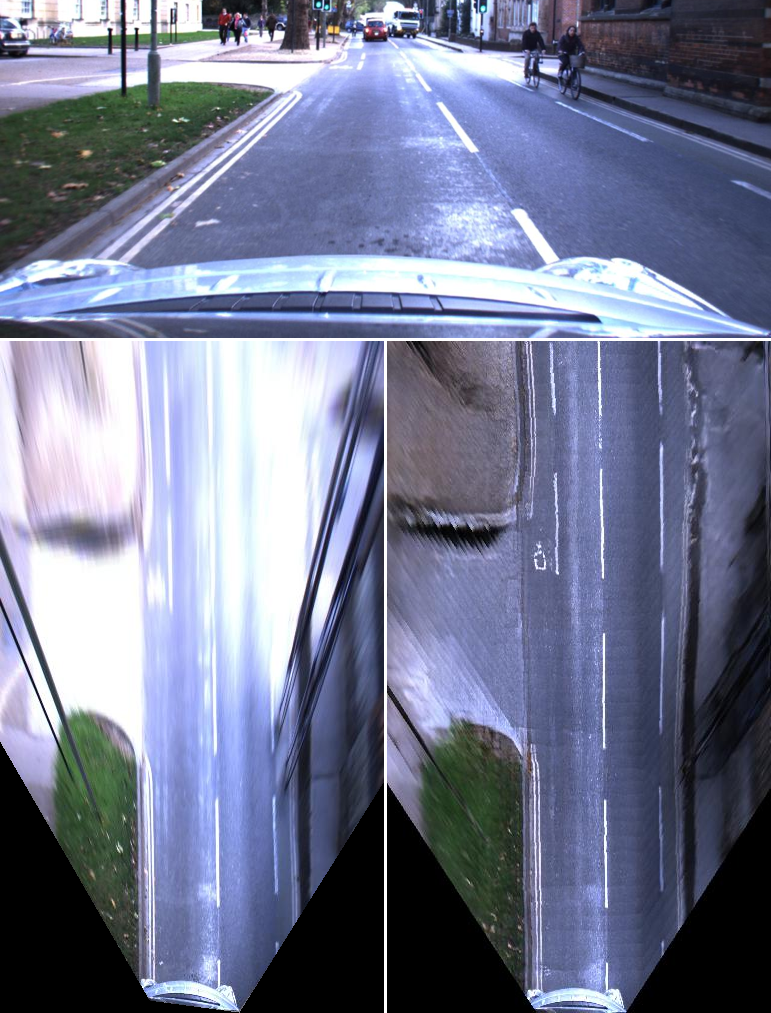}
	\caption{Boosted Inverse Perspective Mapping (IPM) to improve the understanding of road scenes.
	\emph{Left:} Top-down view created by applying a homography-based IPM to the front-facing image (\emph{top}), leading to unnatural blurring and stretching of objects at further distance. \emph{Right:} Improved top-down view generated by our Incremental Spatial Transformer GAN, containing sharper features and a homogeneous illumination, while dynamic objects (i.e. the two cyclists) are automatically removed from the scene.
	}
	\label{fig:introfig}
\end{figure}

In practice, IPM works well in the immediate proximity of the vehicle (assuming the road surface is planar).
However, the geometric properties of objects in the distance are affected unnaturally by this non-homogeneous mapping, as shown in Fig.~\ref{fig:introfig}.
This limits the performance of applications in terms of their accuracy and the distance at which they can be applied reliably.
More crucial, however, is the effect of inaccurate mappings on the semantic interpretation of scenes, where small inaccuracies can lead to significant qualitative differences.
As we demonstrate in Section~\ref{sec:sceneunderstanding} (Table~\ref{tab:scenegraph}), these qualitative differences can manifest themselves in many ways, including missing lanes and/or late detection of stop lines (or other critical road markings).

To overcome these challenges, we present an adversarial learning approach which produces a significantly improved IPM in real time from a single front-facing camera image.
This is a difficult problem which is not solved by existing methods, due to the large difference in appearance between the frontal view and IPM.
State-of-the-art approaches for cross-domain image translation tasks train (conditional) Generative Adversarial Networks (GANs) to transform images to a new domain~\cite{isola2017image, zhu2017unpaired}.
However, these methods are designed to perform aligned appearance transformations and struggle when views change drastically~\cite{zhu2018generative}.
The latter work, in which a synthetic dataset with \emph{perfect} ground-truth labels is used to learn IPM, is closest to ours.



We demonstrate in this paper that we are able to generate reliable, improved IPM for larger scenes than in~\cite{zhu2018generative}, which are therefore able to directly aid scene understanding tasks.
We achieve this in real time using real-world data collected under different conditions with a single front-facing camera.
Consequently, we must deal with \emph{imperfect} training labels (see Section~\ref{sec:trainingdata}) created from a sequence of images and ego-motion.
An Incremental Spatial Transformer GAN is introduced to address the significant appearance change between the frontal view and IPM.
Compared to analytic IPM approaches our learned model is (1) more realistic with sharper contours at long distance, (2) invariant to extreme illumination under different conditions, and (3) removes dynamic objects from the scene to recover the underlying road layout.
We make the following contributions in this paper:
\begin{itemize}
	\item we introduce an Incremental Spatial Transformer GAN for generating boosted IPM in real time;
	\item we explain how to create a dataset for training IPM methods on real-world images under different conditions; and
	\item we demonstrate that our boosted IPM approach improves the detection of road markings as well as the semantic   interpretation of road scenes in the presence of occlusions and/or extreme illumination.
\end{itemize}

\section{Related Work}
\ignore{\tom{shorten this?}
\pmnnotes{yes, defo dont talk about SLAM. Spend more space on results if you can. Really don't want reviewers saying this is a fire and forget paper - as in "seems to work alright"}}

\subsubsection*{\textbf{Improved IPM}}
As indicated in Section \ref{sec:introduction}, many applications can be found in the literature that apply IPM.
They rely on three assumptions: (1) the camera is in a fixed position with respect to the road, (2) the road surface is planar, and (3) the road surface is free of obstacles.
Remarkably, relatively few approaches exist that aim to improve inaccurate IPM, in case one or more of these assumptions are not satisfied.

Several works have tried to adjust for inaccuracies caused by invalidity of the first two assumptions. The authors of \cite{nieto2007stabilization, zhang2014robust} used vanishing point detection, \cite{bertozzi1998extension} estimated the slope of the road according to the lane markings, and \cite{jeong2016adaptive} employed motion estimation obtained from SLAM.
Invalidity of the third assumption is tackled in \cite{oliveira2015multimodal} by using a laser scanner to exclude obstacles from being transformed to IPM.
Another approach \cite{lin2012vision, cerri2005free, menendez2017new} creates a look up table for all pixels, by taking into account the distance of objects on the road surface, in order to reduce artefacts at further distance. However, these methods generally assume simple environments (i.e. highway). Contrarily, we learn a non-linear mapping more suited for urban scenes.



Very recently, \cite{zhu2018generative} proposed the first learning approach for IPM using a synthetic dataset. The authors introduced BridgeGAN which employs the homography IPM to bridge the significant appearance gap between the frontal view and bird's-eye view.
In contrast, we use real-world data and consequently \emph{imperfect} labels to generate boosted IPM for larger scenes. Therefore, our learned mapping is directly beneficial for scene understanding tasks (see Section~\ref{sec:scenegraph}).

\subsubsection*{\textbf{Semantic IPM}}
Several methods use the semantic relations between the two views for different tasks.
In \cite{sengupta2012automatic, mattyus2016hd} conditional random fields in the frontal view and IPM are optimized to retrieve a coarse semantic bird's-eye-view map from a sequence of camera images.
A joint optimization net is trained in \cite{zhai2017predicting, regmi2018cross} to align the semantic cues of the two views. The authors then train a GAN to synthesize a ground-level panorama from the coarse semantic segmentation. However, because aerial images differ significantly in appearance from the ground view, there is a lack of texture and detail in the synthesized images.
We generate a more detailed IPM by learning a direct mapping of the pixels from the frontal view which is more useful for autonomous driving applications.


\subsubsection*{\textbf{GANs for Novel View Synthesis}}
The rise of GANs has made it possible to generate new, realistic images from a learned distribution.
In order to guide the generation process towards a desired output, GANs can be conditioned on an input image \cite{isola2017image, wang2018highresolution}.
Until now, these methods were restricted to perform aligned appearance transformations.


In~\cite{jaderberg2015spatial}, the spatial transformer module was introduced to learn transformations of the input to improve classification tasks.
The authors of~\cite{yan2016perspective, zhou2016view} used similar ideas to synthesize new views of 3D objects or scenes.
More recently, these two fields were combined in~\cite{rezende2016unsupervised, azadi2018compositional}. In the latter work, realistic compositions of objects are generated for a new viewpoint. However, these techniques are limited to toy datasets or distort real-world scenes with dynamic objects.






\section{Boosted IPM using an Incremental Spatial Transformer GAN}

\label{sec:network}

\subsection{Network Overview}

As a starting point, we use a state-of-the-art architecture similar to the global enhancer of~\cite{wang2018highresolution}, without employing boundary or instance maps.
Additionally, as we expect a slight change in scale from the homography-based IPM image to the stitched training labels (see Section~\ref{sec:trainingdata}), we refrain from using any pixel-wise losses and instead use multi-scale discriminator losses~\cite{wang2018highresolution} combined with a perceptual loss~\cite{johnson2016perceptual, dosovitskiy2016generating} based on VGG16~\cite{simonyan2014very}.
While VGG16 is trained on the ImageNet~\cite{krizhevsky2012imagenet} dataset, thus being more suitable for frontal rather than bird's-eye-view images of road scenes, we still leverage the stability of its encoded features in this study.
Retraining VGG16 on bird's-eye-view images of road scenes or swapping it out for a more suitable model, may improve the quality of the generated images, but this is beyond the scope of this study.

\begin{figure}[!t]
	\centering
	\includegraphics[width=\columnwidth]{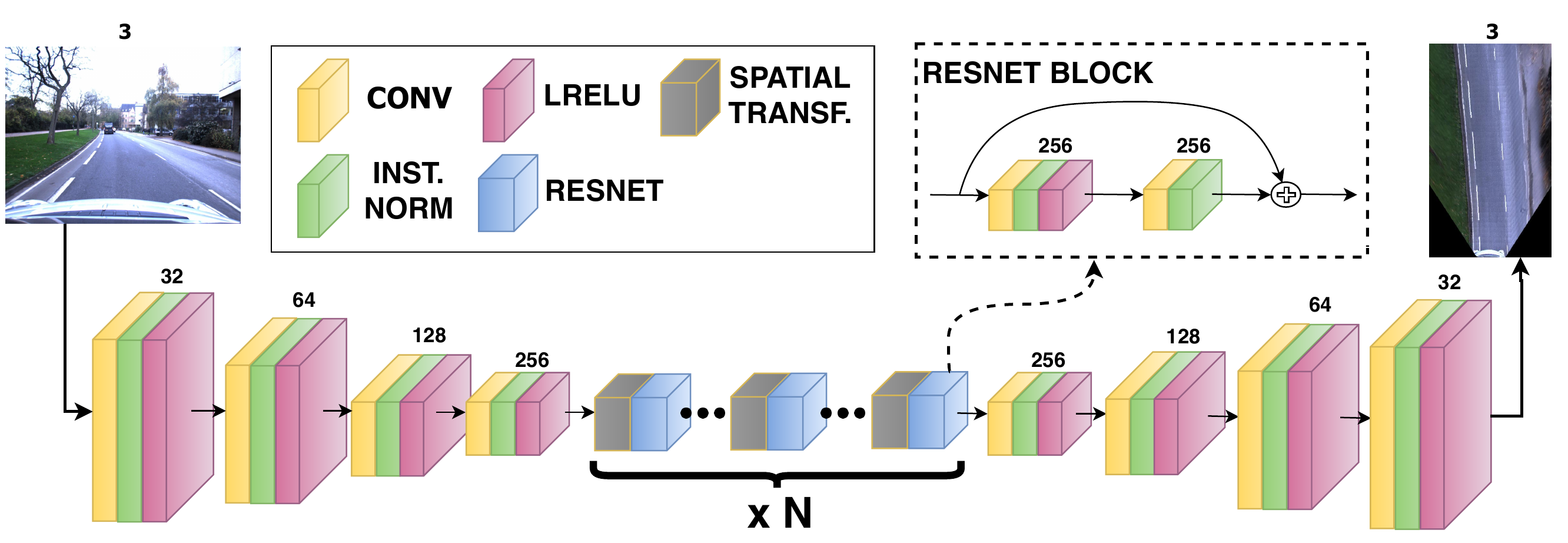}
	\caption{The architecture of the generator of the network. The bottleneck of the model contains a series of $N$ sequential blocks. Each block performs an incremental perspective transformation of $n$ degrees, so that the bottleneck as a whole transforms the features from frontal to bird's-eye view. After every transformation, the features are sharpened by a ResNet block before the next transformation is applied. This process is depicted in more detail in Fig. \ref{fig:incremental}.}
	\label{fig:network}
\end{figure}

Our model follows a largely traditional downsample-bottleneck-upsample architecture, where we reformulate the bottleneck portion of the model as a series of $N_{\mathrm{STRes}}$ blocks that perform incremental perspective transformations followed by feature enhancement.
Each block contains a Spatial Transformer (ST)~\cite{jaderberg2015spatial} followed by a ResNet layer~\cite{he2016deep}.
The structure of the generator is presented in Fig.~\ref{fig:network}.
For an in-depth description of the remaining architecture, the reader is directed towards the paper and supplemental material of~\cite{wang2018highresolution}.

\subsection{Spatial ResNet Transformer}

\begin{figure}[!t]
	\centering
	\includegraphics[width=0.75\columnwidth]{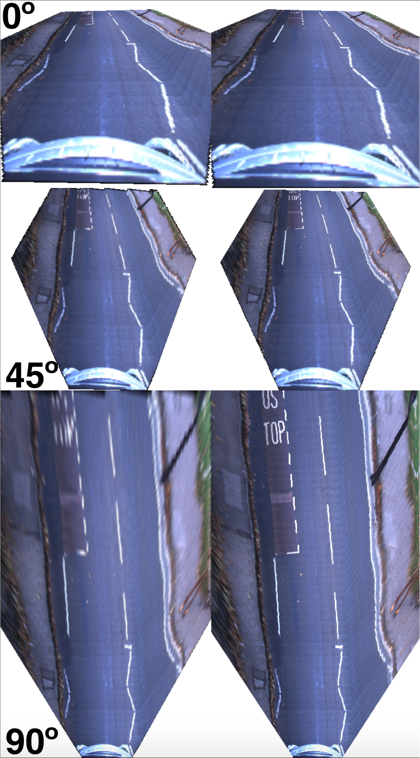}
	\caption{Conceptual visualization of the sequential incremental transformations (i.e. $N=3$, from $0\degree$ to $90\degree$ degrees down the rows) occurring in the bottleneck of the generator. The left column shows the features immediately after the transformation is applied, consequently they are stretched and blurred (e.g. BUS STOP letters). The right column shows how the ResNet blocks learns to sharpen these features to create the improved IPM before the next transformation is applied. Note that in reality the bottleneck has 512 feature maps instead of the 3 RGB channels depicted here for demonstration purposes.}
	\label{fig:incremental}
\end{figure}

Since far-away real-world features are represented by a smaller pixel area as compared to identical close-by features, a direct consequence of applying a full perspective transformation to the input is increased unnatural blurring and stretching of the features at further distance.
To counteract this effect, our model divides the full perspective transformation into a series of $N_{\mathrm{STRes}}$ smaller incremental perspective transformations, each followed by a refinement of the transformed feature space using a ResNet block~\cite{he2016deep}.
The intuition behind this is that the slight blurring that occurs as a result of each perspective transformation is restored by the ResNet block that follows it, as conceptually visualized in Fig.~\ref{fig:incremental}.
To maintain the ability to train our model end-to-end, we apply these incremental transforms using Spatial Transformers~\cite{jaderberg2015spatial}.

Intuitively, a Spatial Transformer is a mechanism, which can be integrated in a deep-learning pipeline, that warps an image using a parametrization (e.g. an affine or homography transformation matrix) conditioned on a specific input signal.
Formally, each incremental spatial transformer is an end-to-end differentiable sampler, represented in our case by two major components:

\begin{itemize}
    \item a convolutional network which receives an input $I$ of size $H_{I}*W_{I}*C$, where $H_{I}$, $W_{I}$ and $C$ represent the height, width, and number of channels of the input respectively, and outputs a parametrization $M_{\mathrm{loc}}$ of a perspective transformation of size $3*3$, and;
    \item a Grid Sampler which takes $I$ and $M_{\mathrm{loc}}$ as inputs, creates a mapping matrix $M_{\mathrm{map}}$ of size $H_{O}*W_{O}*2$, where $H_{O}$ and $W_{O}$ represent the height and width of the output $O$. $M_{\mathrm{map}}$ maps homogeneous coordinates $[x,y,1]^T$ to their new warped position given by $M_{\mathrm{loc}}*[x,y,1]^T$. Finally, $M_{\mathrm{map}}$ is used to construct $O$ in the following way: $O(x,y) = I(M_{\mathrm{map}}(x,y,1),M_{\mathrm{map}}(x,y,2))$.
\end{itemize}

In practice, it is non-trivial to train a spatial transformer (and even less trivial; a sequence of spatial transformers) on inputs with a large degree of self-similarity, such as road scenes.
To stabilize the training procedure, for each incremental spatial transformer, we decompose $M_{\mathrm{loc}}= M_{\mathrm{locref}} * M_{\mathrm{locpert}}$, where $M_{\mathrm{locref}}$ is initialized with an approximate parametrization of the desired incremental homography, and $M_{\mathrm{locpert}}$ is the actual output of the convolutional network and represents a learned perturbation or refinement of $M_{\mathrm{locref}}$.

\subsection{Losses}
Our architecture stems from~\cite{wang2018highresolution}, but does not make use of any instance maps.
Due to the potential misalignment between the output of the network and the labels (see Section \ref{sec:trainingdata}), we rely on a multi-scale discriminator loss and a perceptual loss based on VGG16.
With a generator $G$, $k^{th}$ scale discriminator $D_k$, and $\mathcal{L}_{\mathrm{GAN}}(G,D_{k})$ being the traditional GAN loss defined over $k=3$ scales as in~\cite{wang2018highresolution}, the final objective thus becomes:
\begin{equation}
\begin{split}
   \mathcal{L}_{\mathrm{tot}} = \min_{G} ( ( \max_{D_1,D_2,D_3}\sum_{k=1,2,3}^{}\mathcal{L}_{\mathrm{GAN}}(G,D_{k}) )  + \\ \lambda_{\mathrm{FM}}\sum_{k=1,2,3}^{}\mathcal{L}_{\mathrm{FM}}(G,D_{k})+\lambda_{\mathrm{VGG}}\mathcal{L}_{\mathrm{VGG}}(G)),
    \end{split}
\end{equation}
where $\mathcal{L}_{\mathrm{FM}}(G,D_{k})$ is the multi-scale discriminator loss:
\begin{equation}
 \mathcal{L}_{\mathrm{FM}(G,D_{k})}=\sum_{i=1}^{l_{D}}\frac{1}{w_{i}}{\lVert D_{k}(I_{\mathrm{label}})_{i} - D_{k}(G(I_{\mathrm{input}}))_{i} \rVert}_{1},
\end{equation}
and $\mathcal{L}_{\mathrm{VGG}}(G)$ is the perceptual loss:
\begin{equation}
 \mathcal{L}_{\mathrm{VGG}(G)}=\sum_{i=1}^{l_{P}}\frac{1}{w_{i}}{\lVert \mathrm{VGG}(I_{\mathrm{label}})_{i} - \mathrm{VGG}(G(I_{\mathrm{input}}))_{i} \rVert}_{1},
\end{equation}
with $l_{D}$ denoting the number of discriminator layers used in the discriminator loss, $l_{P}$ denoting the number of layers from VGG16 that are utilized in the perceptual loss, and $I_\mathrm{input}$ and $I_\mathrm{label}$ being the input and label images, respectively.
The weights $w_i=2^{l-i}$ are used to scale the importance of each layer used in the loss.

\subsection{Implementation details}

We choose $N_{\mathrm{STRes}}=6$, $N_{\mathrm{downsample}}=4$, $N_{\mathrm{upsample}}=4$ and $l_{D}=l_{P}=4$. Furthermore, for training, we employ the Adam solver using a base learning rate set at 0.0002, and a batch size of 1, training for 200 epochs.
For the loss trade-off, we empirically set $\lambda_{\mathrm{FM}}=5$ and $\lambda_{\mathrm{VGG}}=2$.
We train our network using 8416 overcast and 4894 nighttime labels.
At run time, the network performs inference in real time ($\approx 20$\,Hz) using an NVIDIA TITAN X.

\section{Creating Training Data for Boosted IPM}
\label{sec:trainingdata}

\begin{figure}[!t]
	\centering
	\includegraphics[width=\columnwidth]{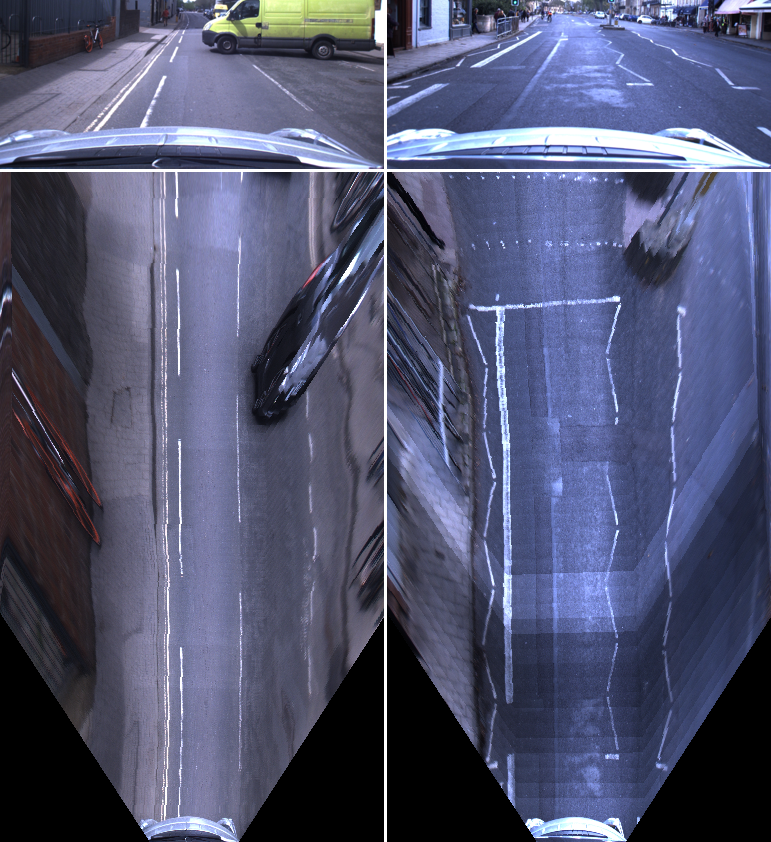}
	\caption{Examples of created training pairs (which show the difficulties of using real-world data) by stitching IPM images generated from future front-facing camera images using the ego-motion obtained from visual odometry. The \emph{left} example illustrates (1) movement of dynamic objects by the time the images are stitched and (2) stretching of objects because they are assumed to be on the road surface. The \emph{right} example shows a significant change of  illumination conditions. \emph{Both} show inaccuracies at further lateral distance (e.g. wavy curb) because of sloping road surface and possibly imprecise motion estimation.
	}
	\label{fig:trainingimages}
\end{figure}

To evaluate our approach, we use the Oxford RobotCar Dataset \cite{maddern20171}, which features a 10-km route through urban environments under different weather and lighting conditions.

In order to create training labels which are a better representation of the real world than the standard, homography-based IPM, we use a sequence of images from the front-facing camera and corresponding visual odometry \cite{WinstonChurchill}, and merge them into a single bird's-eye-view image.

\begin{figure*}[!t]
	\centering
	\includegraphics[width=\textwidth]{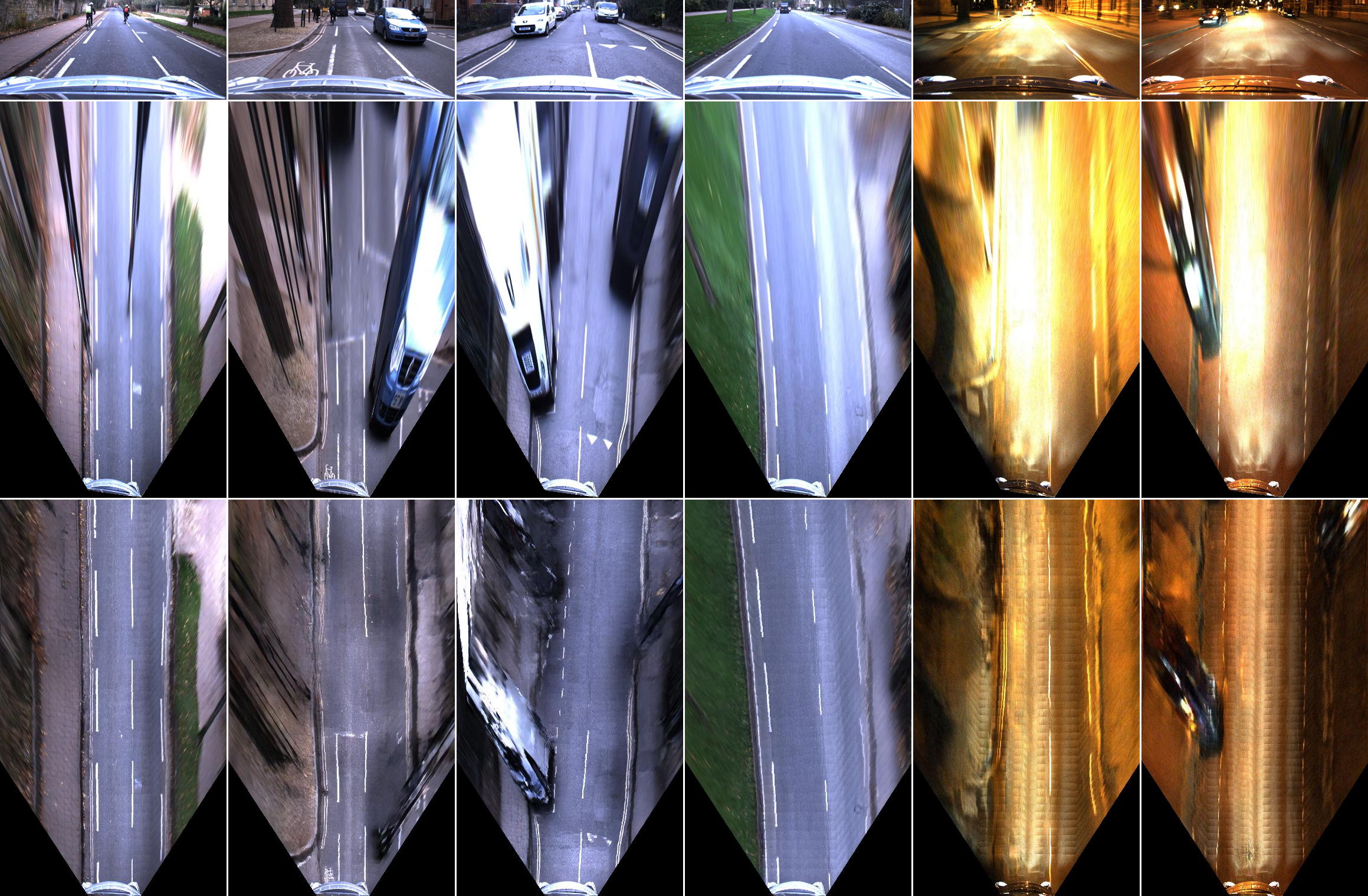}
	\caption{Boosted IPM generated by the network (\emph{bottom}) under different conditions compared to traditional IPM generated by applying a homography (\emph{middle}) to the front-facing camera image (\emph{top}). The boosted bird’s-eye-view images contain sharper features (e.g. road markings), more homogeneous illumination, and automatically remove (dynamic) objects from the scene. Consequently, we infer the underlying road layout, which is directly beneficial for various tasks performed by autonomous vehicles.
	}
	\label{fig:qualitative}
\end{figure*}

From the sensor calibrations and the camera's intrinsic parameters, we compute the transformation which defines the one-to-one mapping between the pixels of the front-facing camera and the bird's-eye view. Then, using the relative transform obtained by visual odometry between the current image frame of the sequence and the initial frame, we stitch the respective pixels of the current frame into the IPM image at the correct pixel positions. This operation is performed iteratively, overwriting previous IPM pixels with more accurate pixels of subsequent frames, until the vehicle has reached the end of its field of view of the initial image.

As the training labels are created from real-world data (in contrast to the synthetic data of \cite{zhu2018generative}), their quality is limited by several aspects (see examples in Fig. \ref{fig:trainingimages}):

\begin{itemize}
    \item Minor inaccuracies in the estimation of the rotation of the vehicle and sloping road surface can lead to imprecise stitching at further lateral distance.
    \item Consecutive image frames may vary significantly in terms of lighting (e.g. due to overexposure), leading to illumination differences in the label which do not naturally occur in the real-world.
    \item Dynamic objects in the front-facing view will appear in a different position in future frames. Consequently, they will appear in unexpected places in the label.
    \item Objects above the road plane (e.g. vehicles, bicyclists, intersection islands, etc.) undergo a large deformation due to the view transformation. We cannot obtain accurate labels for these in real-world scenarios.
\end{itemize}

Due to the aforementioned drawbacks, no direct relation exists between the output (boosted IPM) of our network and the stitched labels. Therefore, it is impossible to incorporate a direct pixel-wise loss function, or employ super-resolution generating networks such as \cite{ledig2017photo}. On the other hand, since we use a sequence of future images, regions that were previously occluded by (dynamic) objects in the initial view are potentially revealed later. This gives the network the ability to learn the underlying road layout irrespective of occlusions or extreme illumination.

\section{Experimental Results}
\label{sec:experiments}
In this section we present qualitative results generated under different conditions.
Due to the nature of the problem, it is extremely hard to capture ground-truth labels in the real world (see Section \ref{sec:trainingdata}), and thus to present quantitative results for our approach.
Furthermore, the synthetic dataset used in \cite{zhu2018generative} is not publicly available.
However, we demonstrate that our boosted IPM has a significant qualitative effect on the semantic interpretation of real-world scenes.
Lastly, we show some limitations of the presented framework.
\pmnnotes{somewhere we need to say why a quantitative set of results are hard to acquire}

\subsection{Qualitative Evaluation}
Fig.~\ref{fig:qualitative} shows qualitative results on a RobotCar test dataset. The results demonstrate that the network has learned the underlying road layout of various urban traffic scenarios. Semantic road features such as parking boxes (i.e. small separators) and stop lines are inferred correctly. Furthermore, dynamic objects, which occlude parts of the scene, are removed and replaced by the correct road/lane boundaries, making the representation more suitable for scene understanding and planning. The boosted IPM contains sharper road markings, which improves the performance of tasks such as lane detection. Lastly, the new view offers a more homogeneous illumination of the road surface, which is beneficial for all tasks that require image processing.

Additionally, we show that our framework is not limited to datasets recorded under overcast conditions. Although artificial lighting during nighttime introduces artefacts in the output, we are still able to significantly improve the representation of the underlying layout of the scene.

\begin{center}
  \begin{table*}
    {\small \addtolength{\tabcolsep}{-5pt}
   \hfill{}
  \caption{Qualitative Effects of IPM Methods on Roadmarking Detection and Scene Interpretation}
  \label{tab:qualitative}
  \hfill{}
      \begin{tabular}{
      >{\centering\arraybackslash}p{0.4\columnwidth} >{\centering\arraybackslash}p{0.365\columnwidth} >{\centering\arraybackslash}p{0.365\columnwidth} >{\centering\arraybackslash}p{0.89\columnwidth}}
     \toprule

     \textbf{Original}  &
     \multicolumn{2}{c}{\textbf{Road marking Detection \cite{bruls2018mark}}}
     & \textbf{Scene Interpretation \cite{kunze2018scene}} \\

     & Homography
     & Boosted IPM
     & (generated from detected road markings)\\
    \midrule

    \begin{overpic}[width=.4\columnwidth]{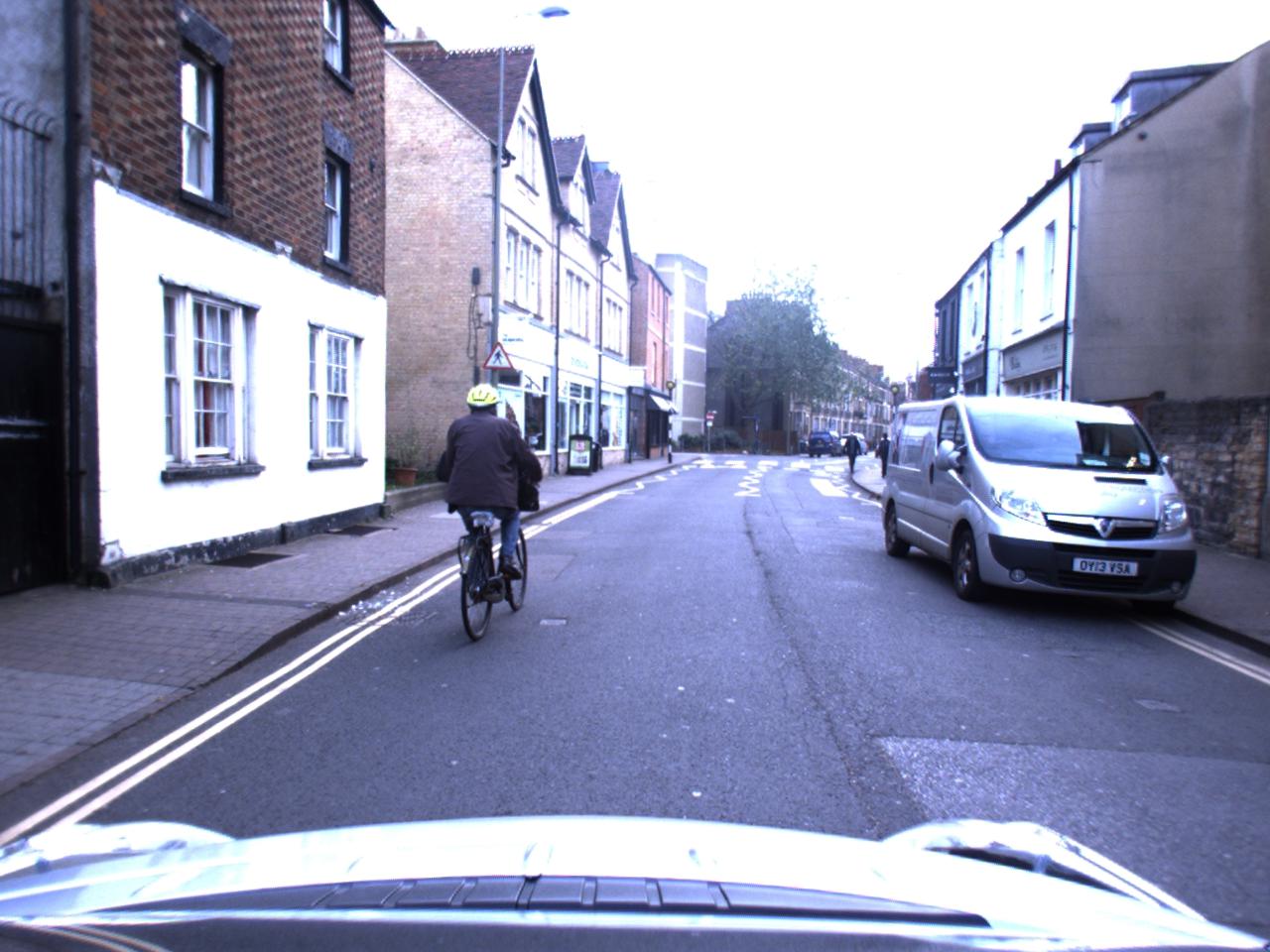}
        \put(5,63){\color{yellow}\Large \bf (A)}
    \end{overpic}
    & \includegraphics[width=.17\columnwidth]{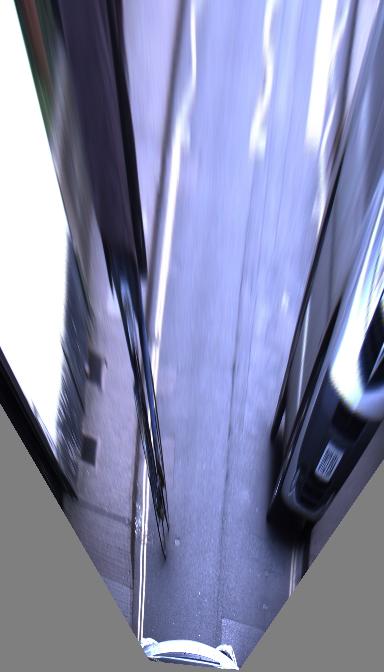}
      \includegraphics[width=.18\columnwidth]{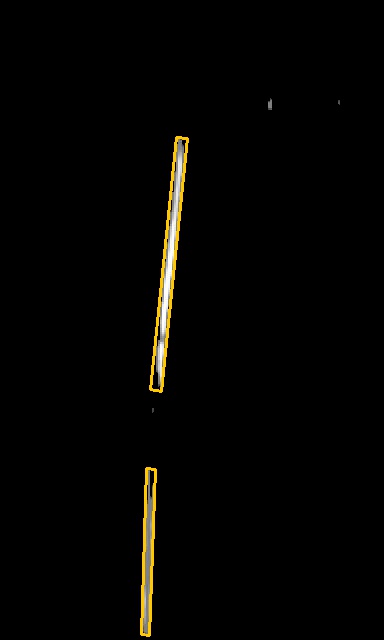}
    & \includegraphics[width=.17\columnwidth]{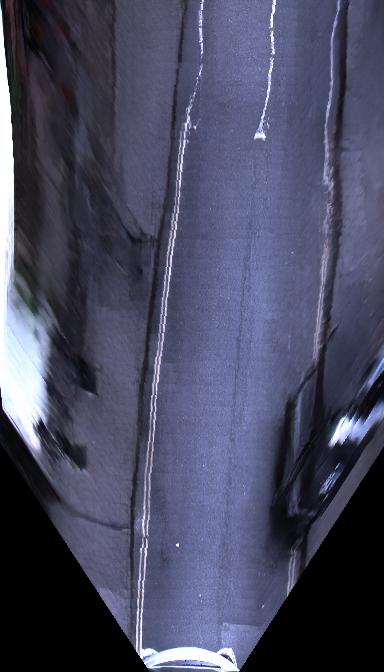}
      \includegraphics[width=.18\columnwidth]{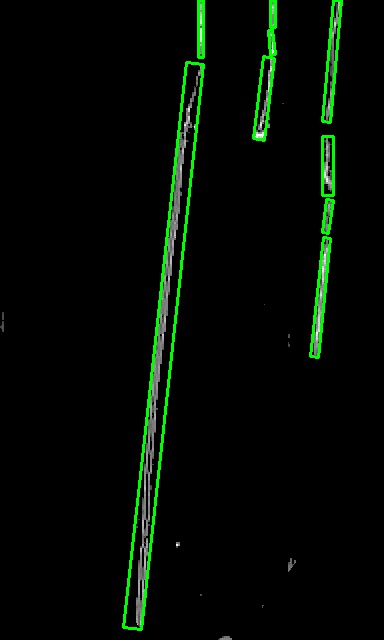}
    & \includegraphics[height=18ex,width=0.9\columnwidth]{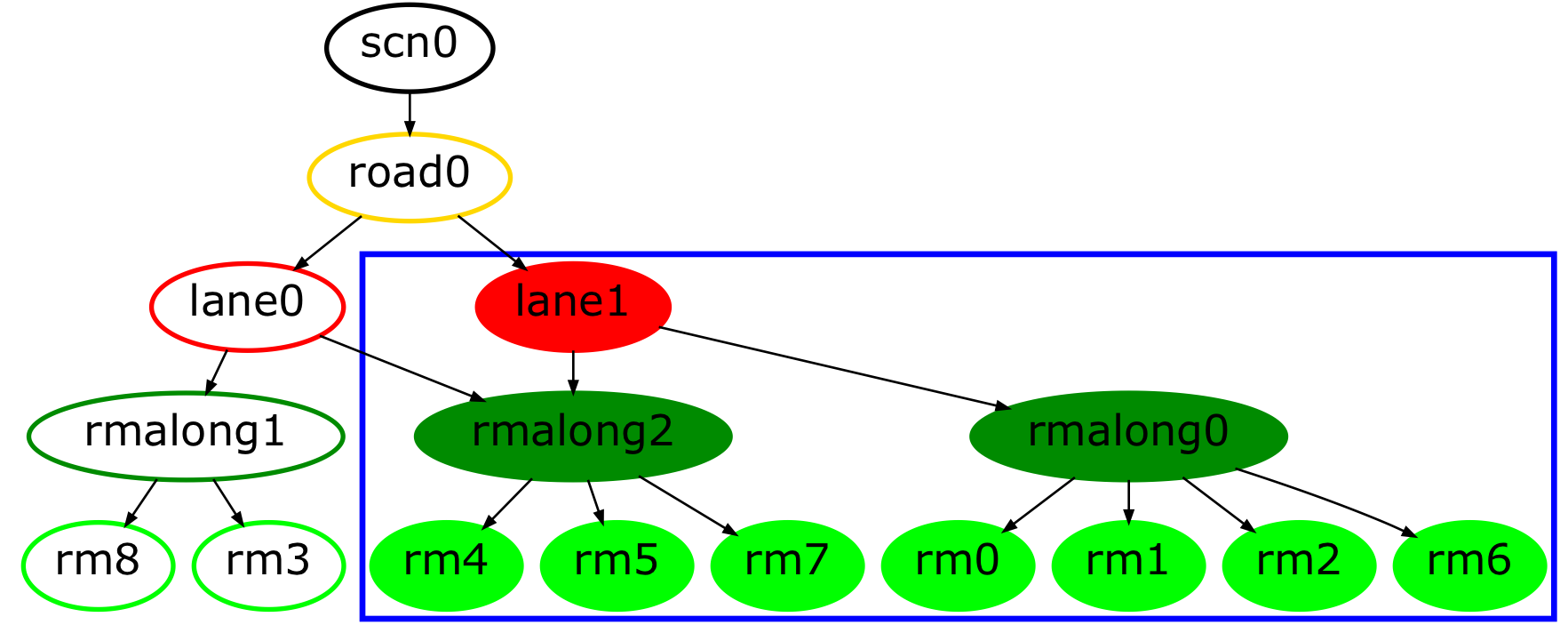}\\ \midrule


    \begin{overpic}[width=.4\columnwidth]{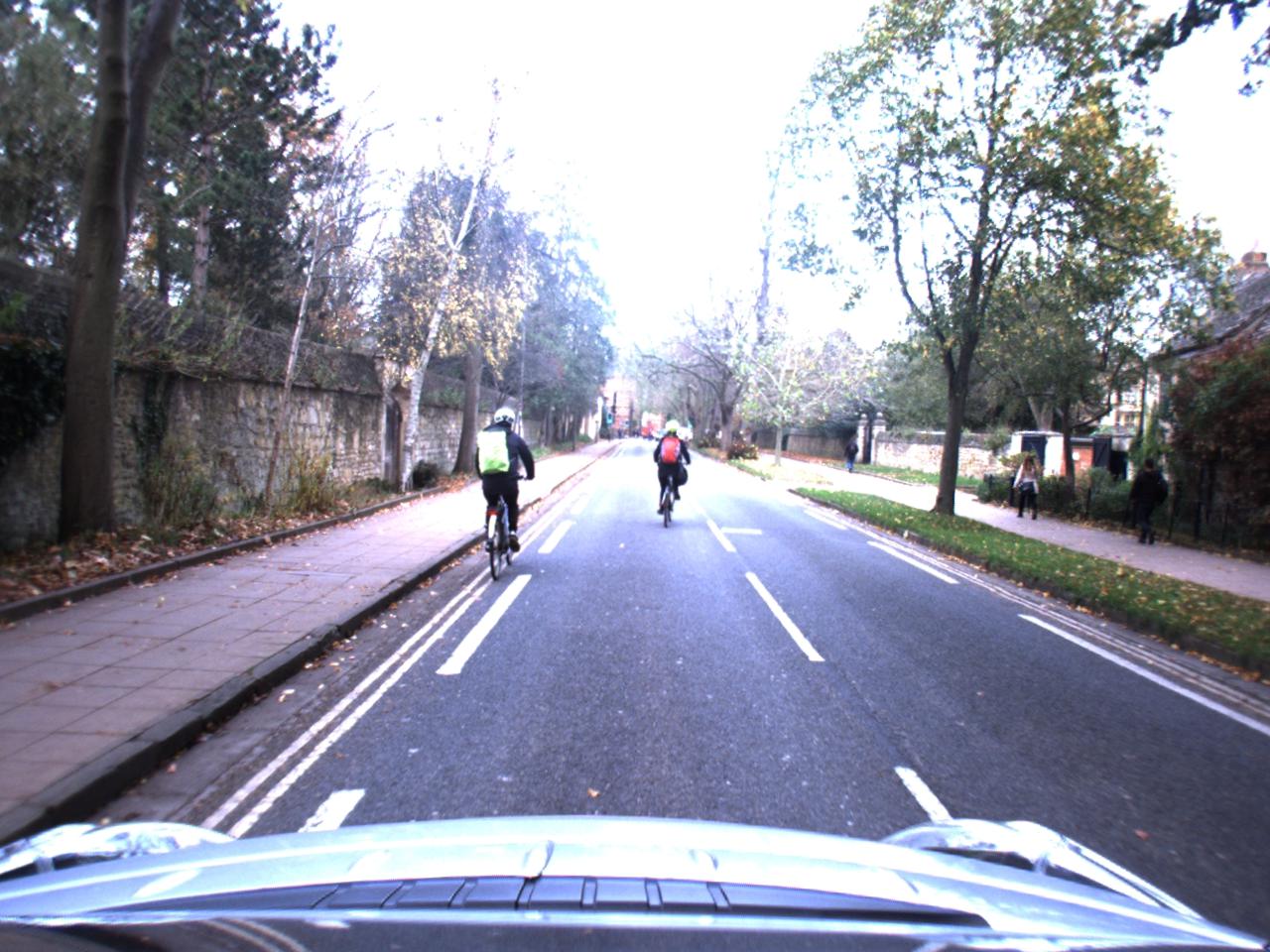}
        \put(5,63){\color{yellow}\Large \bf (B)}
    \end{overpic}
    & \includegraphics[width=.17\columnwidth]{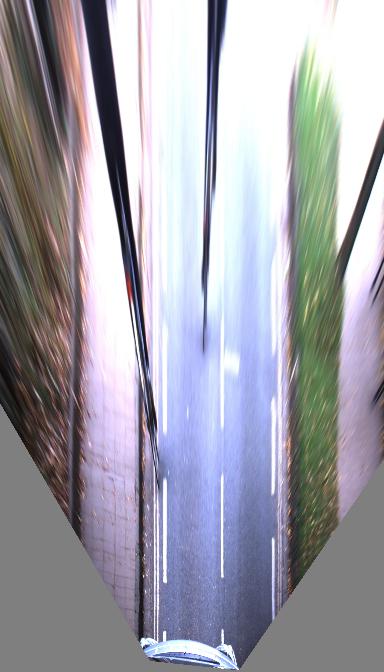}
      \includegraphics[width=.18\columnwidth]{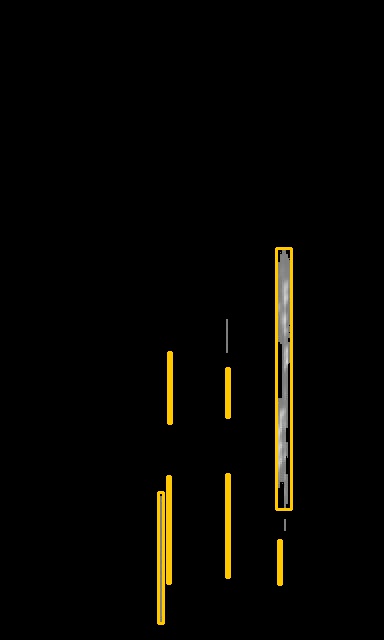}
    & \includegraphics[width=.17\columnwidth]{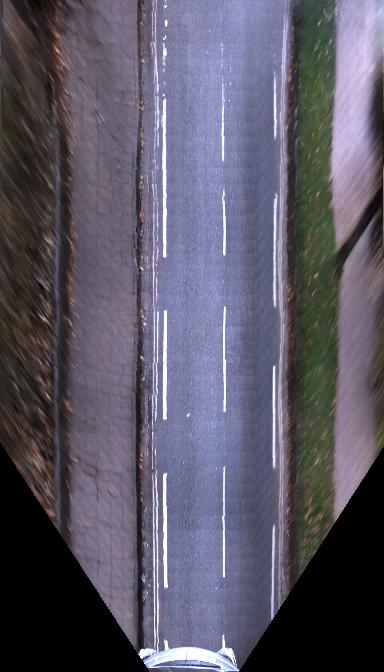}
      \includegraphics[width=.18\columnwidth]{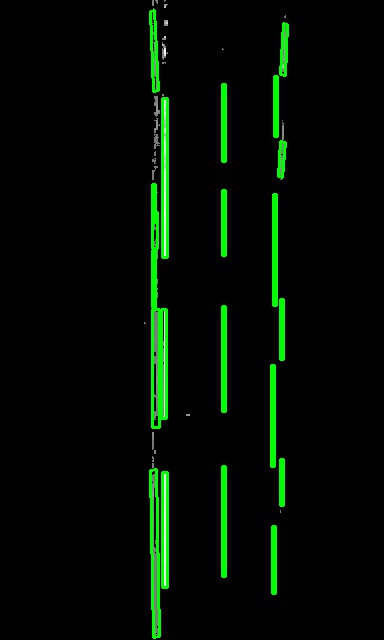}
    & \includegraphics[height=18ex,width=0.9\columnwidth]{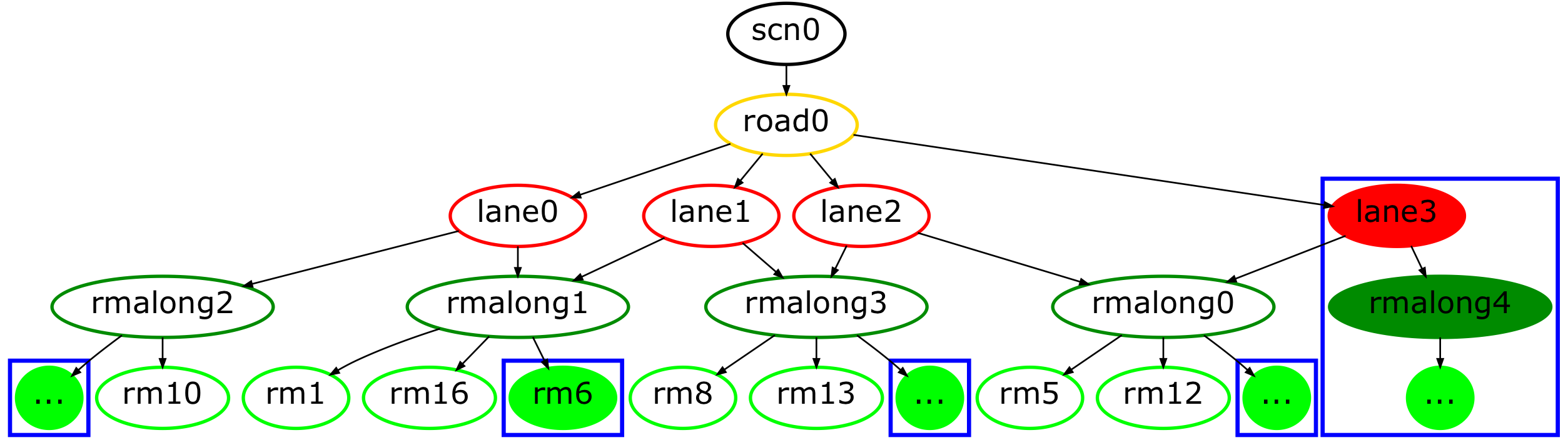}\\ \midrule


    \begin{overpic}[width=.4\columnwidth]{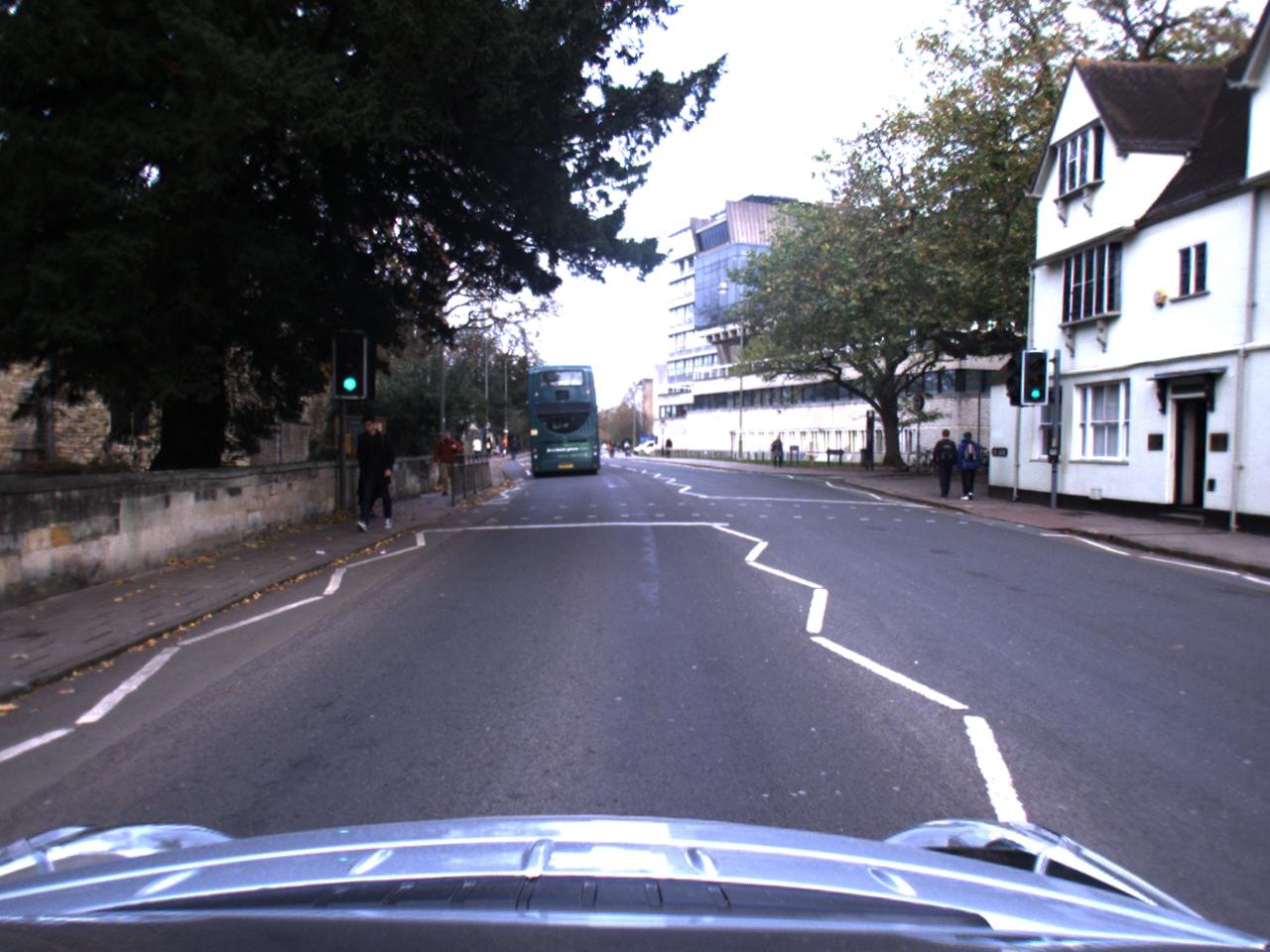}
        \put(5,63){\color{yellow}\Large \bf(C)}
    \end{overpic}
    & \includegraphics[width=.17\columnwidth]{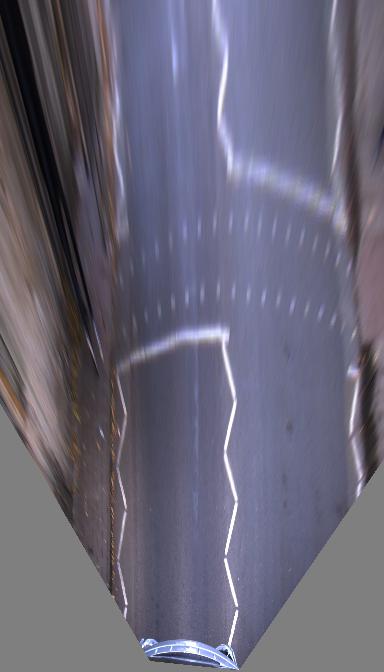}
      \includegraphics[width=.18\columnwidth]{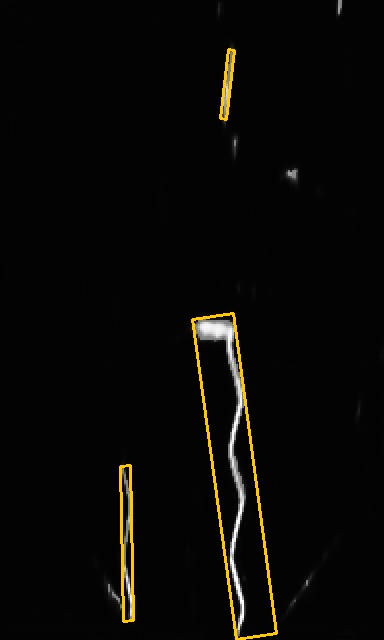}
    & \includegraphics[width=.17\columnwidth]{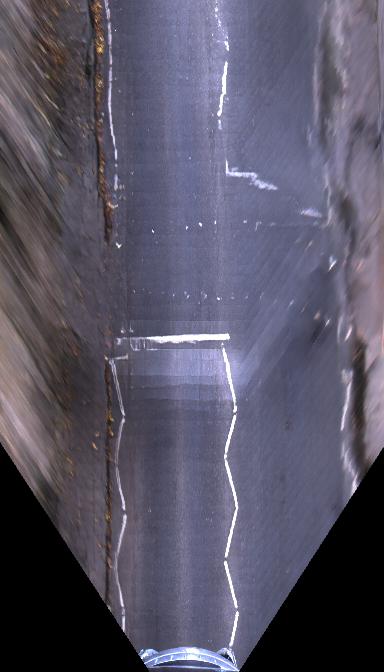}
      \includegraphics[width=.18\columnwidth]{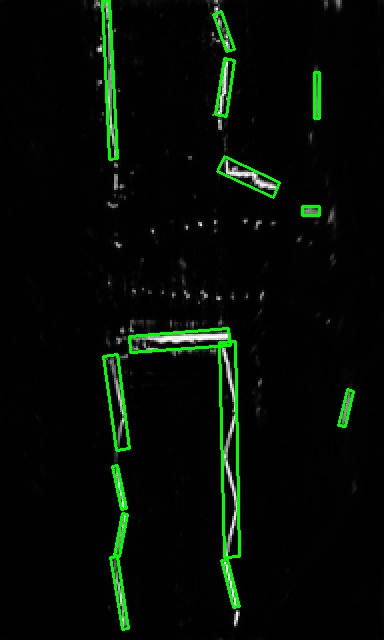}
    & \includegraphics[height=18ex\textbf{},width=0.9\columnwidth]{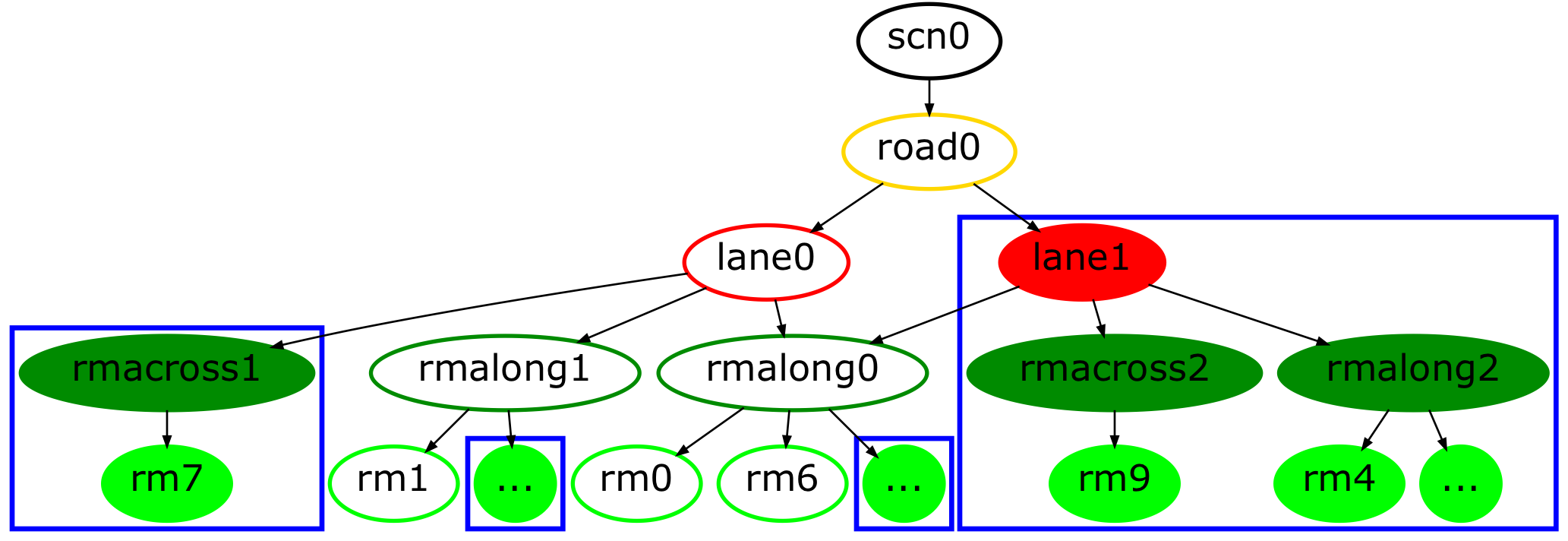}\\ 



    \bottomrule
    \end{tabular}
    \label{tab:scenegraph}
       }
    \hfill{}
  \end{table*}
\end{center}

\subsection{Employing Boosted IPM for Scene Interpretation}
\label{sec:sceneunderstanding}

\label{sec:scenegraph}
We demonstrate the effectiveness of our improved IPM approach for the application of road marking detection~\cite{bruls2018mark} and scene interpretation~\cite{kunze2018scene} (cf. Table~\ref{tab:scenegraph}). Table~\ref{tab:scenegraph} shows the original front-facing camera image, the bird's-eye views (homography-based as well as our boosted IPM) and their corresponding road marking detections, and the generated graph-based scene description.

The input to the scene interpretation process is the binary image mask of the detected road markings.
Within these experiments this input is either provided by the homography-based IPM or by our boosted IPM.
We then cluster the road marking pixels into groups and compute a set of spatial properties and relations.
Based on the spatial information and a learned probabilistic grammar, which captures the road layout of scenes, a hierarchical, graph-based scene description is generated including information about roads, lanes and road markings (which are grounded in image space). The reader is directed towards~\cite{kunze2018scene} for more details.

As the overall scene interpretation is based on the segmentation of road markings, the quality of the road marking detection has a major impact on the generated scene graph, as demonstrated later.
Experimentally, we have verified that boosted IPM allows us to more robustly detect road markings (1) at greater distance and (2) in more detail, and (3) infer road markings occluded by dynamic objects such as cars and cyclists.
These improvements are possible because boosted IPM contains sharper features with more consistent geometric properties (at further distance) and learns the underlying road layout.

We have trained a road marking detection network for each view separately (because we expect a difference in learned features) with an equivalent setup according to \cite{bruls2018mark}.
Labels (in the front-facing image) were generated automatically by using the techniques of \cite{bruls2018mark} and mapped down into IPM to match the input images.
In addition, the boosted IPM road marking labels were stitched similarly to the camera images.
Although the labels are not equivalent to the ground-truth, they have proven to be sufficient for training purposes if regularization techniques are applied.
The increase in performance for road marking detection in the boosted IPM has immediate consequences for the interpretation of scenes. In general, all interpretations (scene graphs) benefit from more accurate road marking detection.
Table~\ref{tab:scenegraph} depicts qualitative differences in the scene graphs\footnote{In the scene graphs, the qualitative differences resulting from our boosted IPM method are indicated by filled nodes grouped in blue boxes.}. In the following we discuss the individual scenes.

\paragraph*{Scene (A)} The vehicle approaches a pedestrian crossing which is signaled by the upcoming zig-zag lines (visible at the top of the image).
While these road markings are visible to the human eye in the homography-based IPM, the trained road marking detection network was not able to detect them because of the stretching and blurring at further distance.
However, our boosted IPM produced a bird's-eye-view image with sharper contours for the zig-zag lines and correct reconstruction of the road markings occluded by the vehicle.
This resulted in an improved scene graph which not only captured the right boundary of the ego lane, but also a previously undetected second lane on the right.
Such qualitative differences have substantial impact on the planning and decision making of the vehicle.

\paragraph*{Scene (B)} The vehicle drives on a road with four lanes --- two inner lanes for vehicles and two outer lanes for cyclists --- and  experiences a sudden change in illumination (from a darker foreground to a brighter background). This is clearly visible in the homography-based IPM and consequently leads to a poor detection of road markings. In contrast, our boosted approach produces a top-down view which inpaints learned semantic cues (i.e. road markings) directly over the overexposed area and also excludes the two cyclists. Hence, the resulting scene graph captures more detail as well as an extra lane which was missed in the segmentation resulting from the standard approach.

\paragraph*{Scene} (C) The vehicle approaches a pedestrian crossing which is indicated by both zig-zag and stop lines. Again, the distorted and blurry image resulting from the homography-based IPM leads to a poor detection of road markings. Our boosted approach has generated a more detailed view which led to better road marking detection including the successful identification of the stop lines. The resulting scene graph based on the homography-based IPM not only misses a lane, but crucially also both stop lines.

Such qualitative differences clearly demonstrate the advantage of our proposed method as they have a direct impact on planning and decision making of autonomous vehicles.
While the detection and interpretation of road markings at a greater distance will enable an autonomous vehicle to adapt its behaviour earlier, the detection of road markings behind moving objects will lead to performance that is more robust and safer even when the scene is partly occluded.

\subsection{Failure Cases}
Under certain conditions, the boosted IPM does not accurately depict all details of the bird's-eye view of the scene.

\begin{figure}[!t]
	\centering
	\includegraphics[width=\columnwidth]{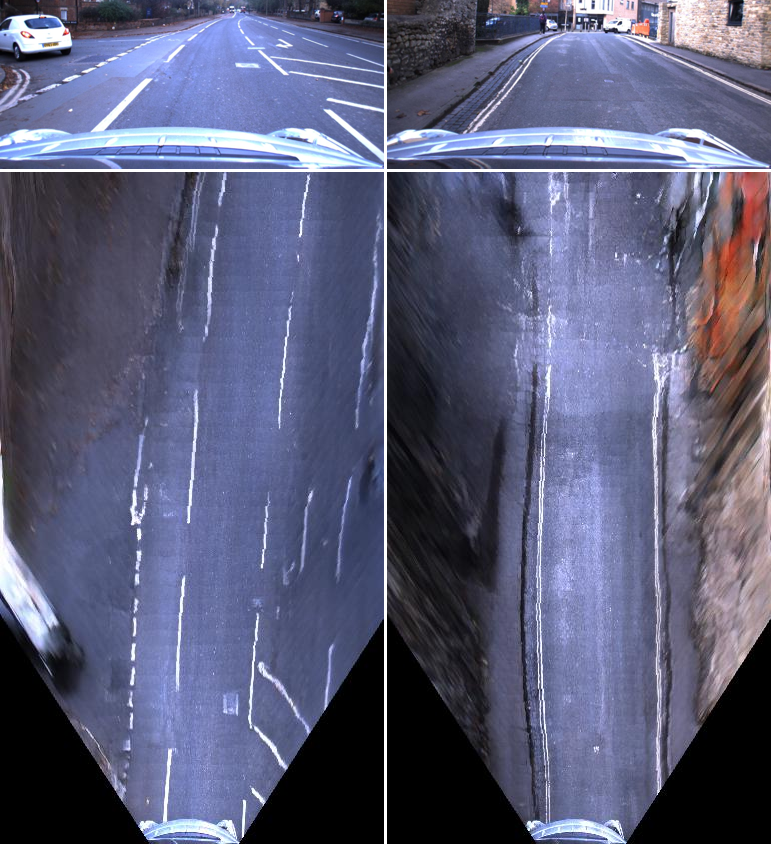}
	\caption{Two cases in which the output of the network does not accurately depict the top-down view of the scene. In the \emph{left} image, the road marking arrow is deformed, because we cannot employ a pixel-wise loss. In the \emph{right} image, the road surface is not flat (sloping upwards), consequently the spatial transformer blocks attempt to map parts of the scene above the horizon, for which the features are not learned.
	}
	\label{fig:failures}
\end{figure}

As we cannot enforce a pixel-wise loss during training (Section \ref{sec:trainingdata}), the shape of certain road markings is not accurately reflected (illustrated in Fig. \ref{fig:failures}). Improvement of the representation of these structural elements will be investigated in future work.

Furthermore, the spatial transformer blocks assume that the road surface is more or less planar (and perpendicular to the $z$-axis of the vehicle). When this assumption is not satisfied, the network is unable to accurately reflect the top-down scene at further distance. This might be solved by providing/learning the rotation of the road surface with respect to the vehicle.

\section{Conclusion}
We have presented an adversarial learning approach for generating boosted IPM from a single front-facing camera image in real time.
The generated results show sharper features and a more homogeneous illumination, while (dynamic) objects are automatically removed from the scene.
Overall, we infer the underlying road layout, which is directly beneficial for tasks performed by autonomous vehicles such as road marking detection, object tracking, and path planning.

In contrast to existing approaches, we used real-world data collected under different conditions, which introduced additional issues due to varying illumination and (dynamic) objects, making it impossible to employ a pixel-wise loss during training.
We have addressed the significant appearance change between the views by introducing an Incremental Spatial Transformer GAN.

We have demonstrated reliable, qualitative results in different environments and under varying lighting conditions.
Furthermore, we have shown that the boosted IPM view allows for improved hierarchical scene understanding.

Consequently, our boosted IPM approach can have a significant impact on a wide range of applications in the context of autonomous driving including scene understanding, navigation, and planning.


\ignore{
\section*{Acknowledgment}
\larsnotes{if you need more space could be left out at time of submission}
We gratefully acknowledge the support of NVIDIA Corporation with the donation of the TITAN Xp GPU used for this research.}


\end{document}